\documentclass{article}
\usepackage{url}
\usepackage[T1]{fontenc}
\usepackage{microtype}
\usepackage{graphicx}

\usepackage{subfigure}
\usepackage{booktabs} 
\usepackage{amsfonts}
\usepackage{mathptmx}
\usepackage{amsmath,amsthm}
\newtheorem*{rem}{Remark} 
\usepackage{hyperref}

\usepackage{etoolbox}

\makeatletter
\patchcmd{\@footnotetext}{\footnotesize}{\scriptsize}{}{}
\makeatother
\usepackage[accepted]{icml2021}

\icmltitlerunning{Non-intrusive Learning of Physics-Informed Spatio-temporal Surrogate for Accelerating Design}

\begin{document}

\twocolumn[
\icmltitle{Non-intrusive Learning of Physics-Informed Spatio-temporal Surrogate for Accelerating Design}




\begin{icmlauthorlist}

\icmlauthor{Sudeepta Mondal}{sm}
\icmlauthor{Soumalya Sarkar}{sm}
\vspace{5mm}
\end{icmlauthorlist}

\icmlaffiliation{sm}{Raytheon Technologies Research Center, 411 Silver Lane, East Hartford, CT 06108}

\icmlcorrespondingauthor{Sudeepta Mondal}{sudeepta.mondal2@rtx.com}

\icmlkeywords{Machine Learning, ICML}

]



\printAffiliationsAndNotice{}  
\setkeys{Gin}{draft = false}
\begin{abstract}
Most practical engineering design problems involve nonlinear spatio-temporal dynamical systems. Multi-physics simulations are often performed to capture the fine spatio-temporal scales which govern the evolution of these systems. However, these simulations are often high-fidelity in nature, and can be computationally very expensive. Hence, generating data from these expensive simulations becomes a bottleneck in an end-to-end engineering design process. Spatio-temporal surrogate modeling of these dynamical systems has been a popular data-driven solution to tackle this computational bottleneck. This is because accurate machine learning models emulating the dynamical systems can be orders of magnitude faster than the actual simulations. However, one key limitation of purely data-driven approaches is their lack of generalizability to inputs outside the training distribution. In this paper, we propose a physics-informed spatio-temporal surrogate modeling (PISTM) framework constrained by the physics of the underlying dynamical system. The framework leverages state-of-the-art advancements in the field of Koopman autoencoders to learn the underlying spatio-temporal dynamics in a non-intrusive manner, coupled with a spatio-temporal surrogate model which predicts the behavior of the Koopman operator in a specified time window for unknown operating conditions. We evaluate our framework on a prototypical fluid flow problem of interest: two-dimensional incompressible flow around a cylinder. 
\end{abstract}

\section{\label{sec: introduction}Introduction}
\begin{figure*}[ht]      
\includegraphics[width= 0.75\textwidth]{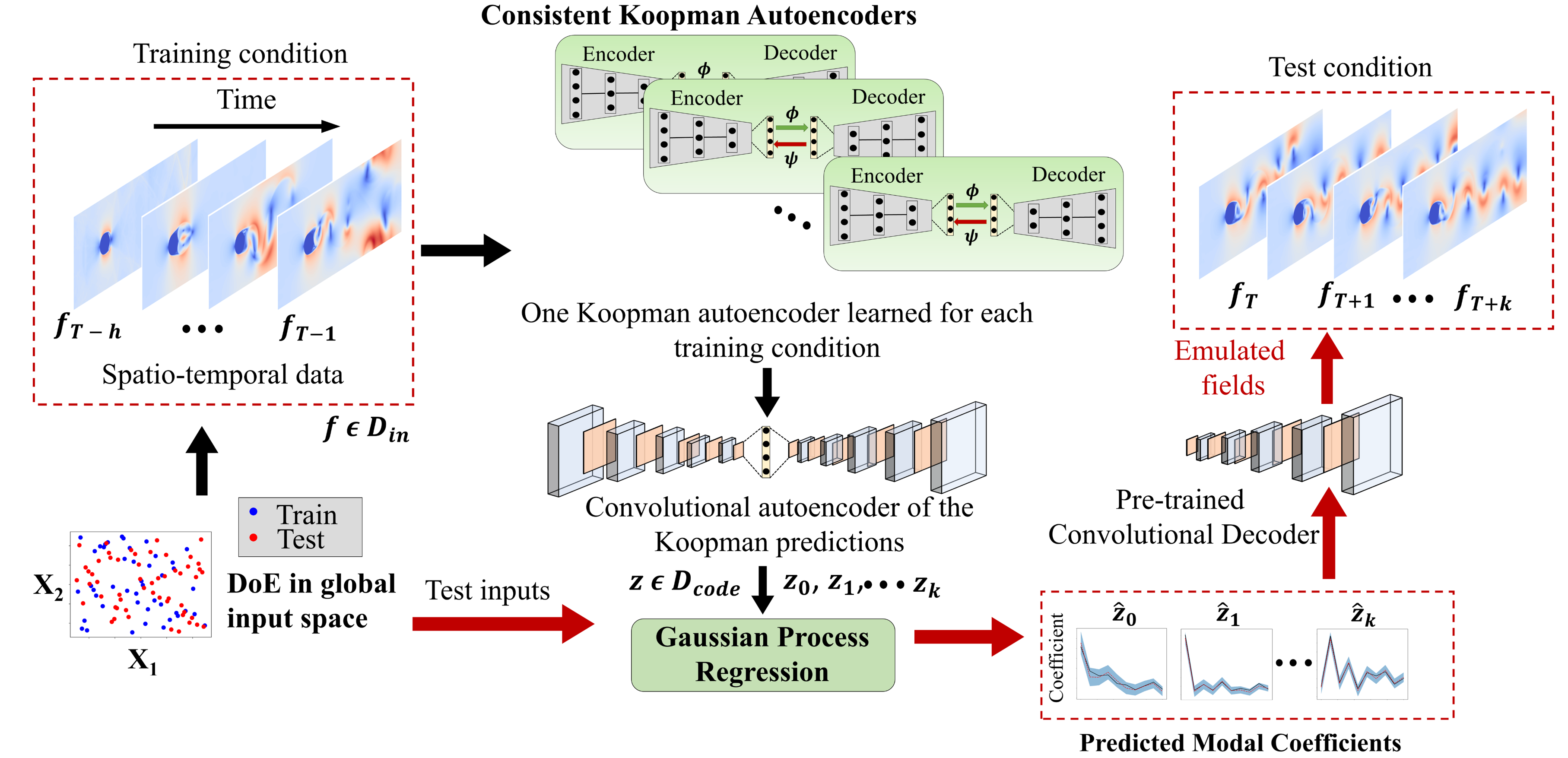} 
\centering
\caption{Schematic overview of the proposed physics-informed spatio-temporal modeling (PISTM) framework of dynamical systems.}
\label{fig:overview} 
\end{figure*}
In the recent years there has been a lot of work involving machine learning (ML) frameworks for emulating spatio-temporally varying fields encountered in nonlinear dynamical systems~\cite{Fukami2021}. These systems are frequently encountered in engineering design, and data from such systems are often generated by high fidelity (HF) simulations that are multi-physics and multi-scale in nature. Such simulations are often computationally very expensive to perform, and become a bottleneck in the overall engineering design process. There has been a lot of work in the domain of data-driven modeling to formulate surrogate models for spatio-temporal dynamical systems. Specifically, deep learning models such as convolutional neural networks (CNN)~\cite{gonzalez_CAE}, long short-term memory networks (LSTM)~\cite{mohan_LSTM}, fully connected neural networks (FCNN)~\cite{kutz_2017, Mondal_SAE_2020} have been extensively used along with more traditional approaches such as proper orthogonal decomposition (POD) and dynamic mode decomposition (DMD)~\cite{Brunton_DMD} for reduced order modeling and emulation of various dynamical systems. 

However, most of these approaches do not incorporate any physical constraints in their formulation, and hence have limitations with respect to their generalizability for input conditions that are outside the training domain. To this end, there has been an ever growing body of work in the field of physics-informed machine learning, whereby ML models are constrained with physical laws in their training phase. \citet{RAISSI2019686} developed a physics-informed neural network (PINN) architecture that is constrained by the Navier Stokes equation to reconstruct the flowfields which have guarantees of satisfying physical laws and boundary constraints. Recently, PINNs have been popularly used in various predictive learning problems involving partial differential equations (PDEs)~\cite{MENG2020109020, souvik_TL, Karniadakis_Review} that govern the dynamics of the systems. However, most PINN formulations~\cite{Karniadakis_Review} require the knowledge of the underlying PDEs governing a physical process, either partially or fully. For any generic dynamical system, where the user is restricted to merely observing the spatio-temporal outputs from the system, the knowledge of the governing equations is often not there. PINN frameworks that rely on the physical laws are not completely suited for such scenarios, and this makes the application of such frameworks challenging for practical engineering applications. 

To this end, researchers have looked into physics-informed approaches which are informed by the physics of dynamical systems in general, and not the underlying physical laws. Some of the notable works in this regard are the ones by~\citet{Erichson_Lyapunov} which incorporate constraints inspired by Lyapunov stability in their framework, and ~\citet{Azencot_koopman, Rice_koopman} which model the underlying dynamical systems using Koopman operators~\cite{Koopman_operator}. Koopman operator theory has been recently receiving considerable attention in the modeling and analysis of several nonlinear systems, with applications in various fields such as fluid mechanics. Koopman's theory assumes that a nonlinear dynamical system can be fully encoded using an infinite dimensional linear operator such that the system evolves linearly in time in the transformed space. Based on this assumption on both forward and backward dynamics of dynamical systems,~\citet{Azencot_koopman, Rice_koopman} showed that consistent Koopman Autoencoders (KAE) are able to perform time series forecasting tasks, along with incorporating high-level constraints in the model. This is a significant advantage when compared to traditional time series prediction techniques in deep learning using Recurrent Neural Network (RNN) architectures such as LSTMs, whereby it is difficult to incorporate such constraints in the model learning phase. However, till date, the research based on these approaches have considered only the temporal learning problem, i.e. forecasting how the dynamical system evolves in time for a given operating and initial condition. The state-of-the-art cannot be directly applicable in practical problems of spatio-temporal surrogate modeling in which it is required to predict how a dynamical system evolves in time for unknown operating and boundary conditions.

In this work, we aim to address this gap using a novel physics-informed spatio-temporal modeling framework (PISTM) which is based on: (a) learning a reduced order model (ROM) of the spatio-temporal evolution predicted by Koopman convolutional autoencoders on available training data (b) employing Gaussian process regression~\cite{Rasmussen2005} models to predict the latent space coefficients of the ROM for the Koopman autoencoder for unknown operating/boundary conditions and (c) using the pre-trained decoder to predict the Koopman evolution for unknown operating/boundary conditions. A schematic overview of the approach is shown in Figure~\ref{fig:overview}. The Koopman autoencoder framework involves fully connected layers for encoding the dynamics of the underlying system, and assume that the dynamics in the latent space evolve linearly both forward and backward in time. This is achieved through the enforcement of the forward mapping $\Phi$ and the backward mapping $\Psi$ to be approximate inverse of each other~\cite{Rice_koopman} (c.f. Figure~\ref{fig:overview}). The ROM of the Koopman autoencoder is a convolutional autoencoder framework, similar in architecture to the one proposed by the authors in a previous work~\cite{mondal2022multifidelity}. Our framework has been validated on the problem of predicting the two-dimensional incompressible viscous flow around a circular cylinder for unknown Reynolds numbers.

\section{Problem Statement}\label{sec: problem_statement}

The problem of interest in this work is to predict the spatio-temporal signature $f(t)$  of a nonlinear dynamical system for a discrete time window of $t = T$ to $t = T + k$ for an unknown operating condition ($\textbf{X}_{test} \in \textbf{X}_{in}$), given that we know how the system behaves for a specified time history of $t = T - h$ to $t = T - 1$ for some operating conditions $\textbf{X}_{train} \in \textbf{X}_{in}$, with $h > 1$. Problems in this setup are very common in many practical engineering design problems, where the challenge is to predict how a complex system would behave in a specified time-frame for an unknown operating condition. It is to be noted that this problem is different and more challenging than classical forecasting problems, whereby given a temporal evolution of the system dynamics for a \textit{particular} condition, it is of interest to predict how the system would evolve in the future timesteps. Koopman autoencoders~\cite{Rice_koopman, Azencot_koopman} have been shown to outperform purely data-driven schemes by incorporating the underlying dynamical systems' constraints in their formulation. However, they are not directly suited for predicting how the system would evolve in time for an unknown operating condition, since each of these models are learned based on a given temporal history for a specified operating condition.
\begin{figure*} [t]\centering
\begin{subfigure}
    \centering
    \includegraphics[width=0.24\textwidth,height=0.24\textheight,keepaspectratio]{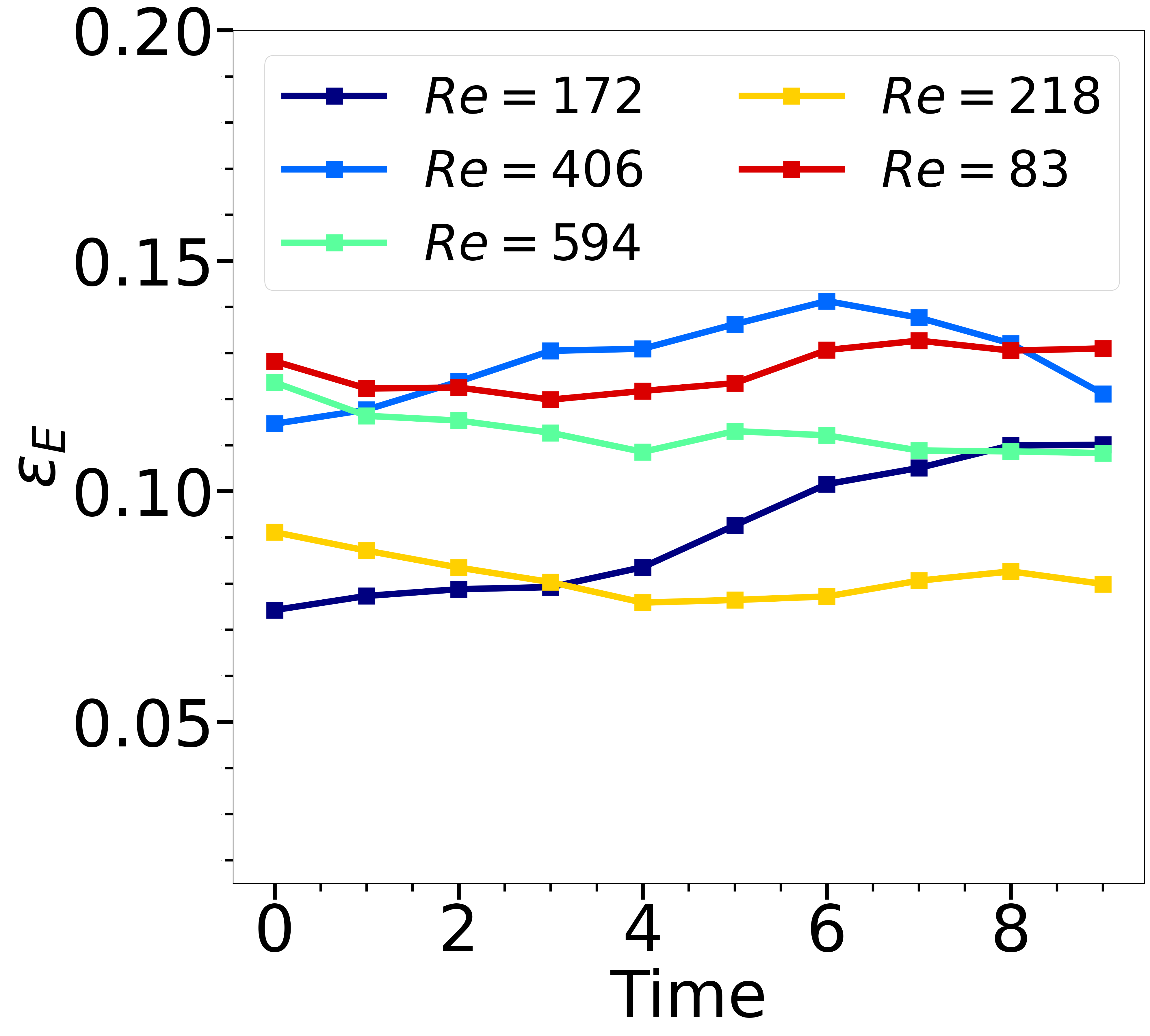}
\end{subfigure}
\begin{subfigure}
    \centering
    \includegraphics[width=0.24\textwidth,height=0.24\textheight,keepaspectratio]{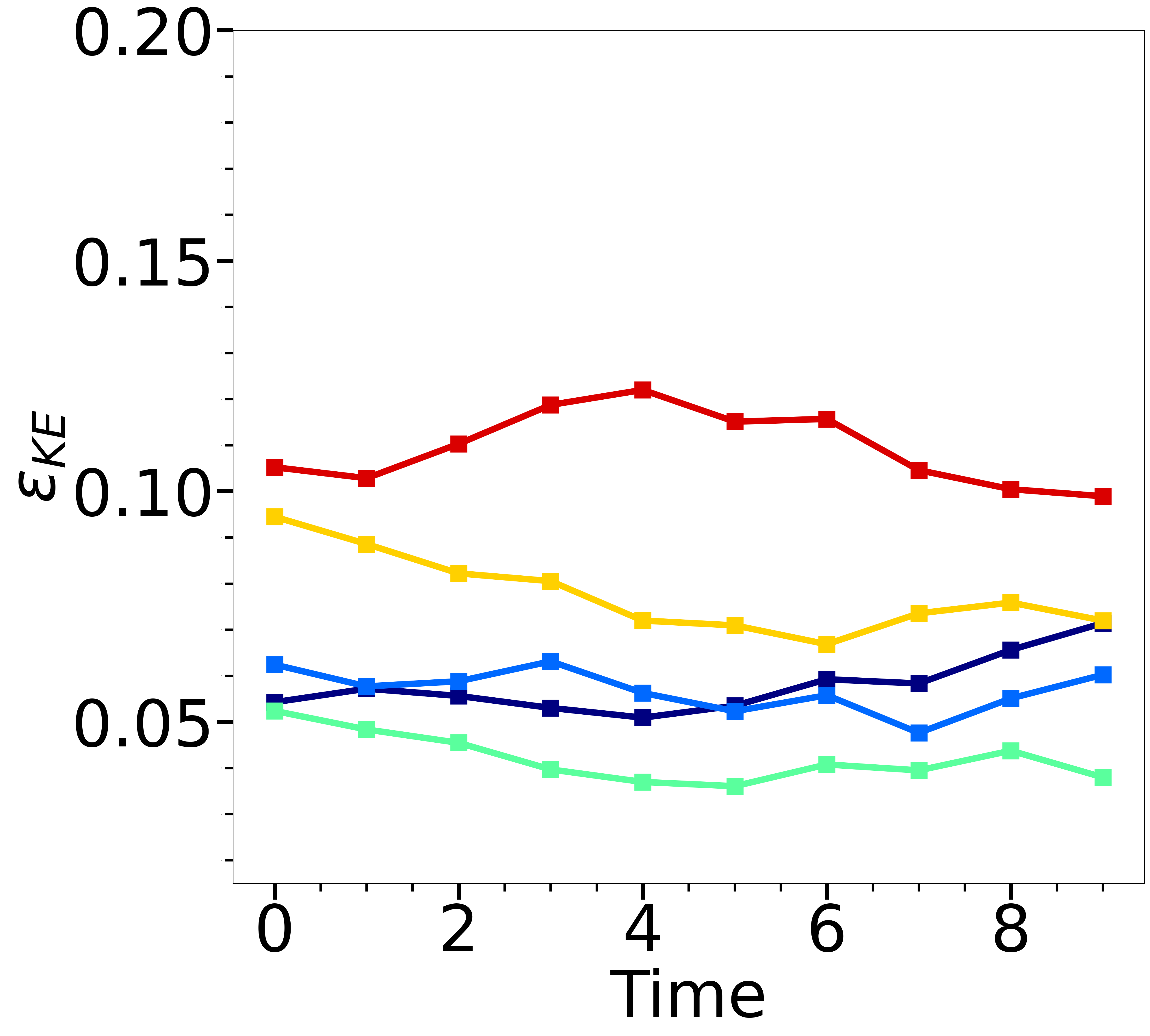}
\end{subfigure}
\begin{subfigure}
    \centering
    \includegraphics[width=0.24\textwidth,height=0.24\textheight,keepaspectratio]{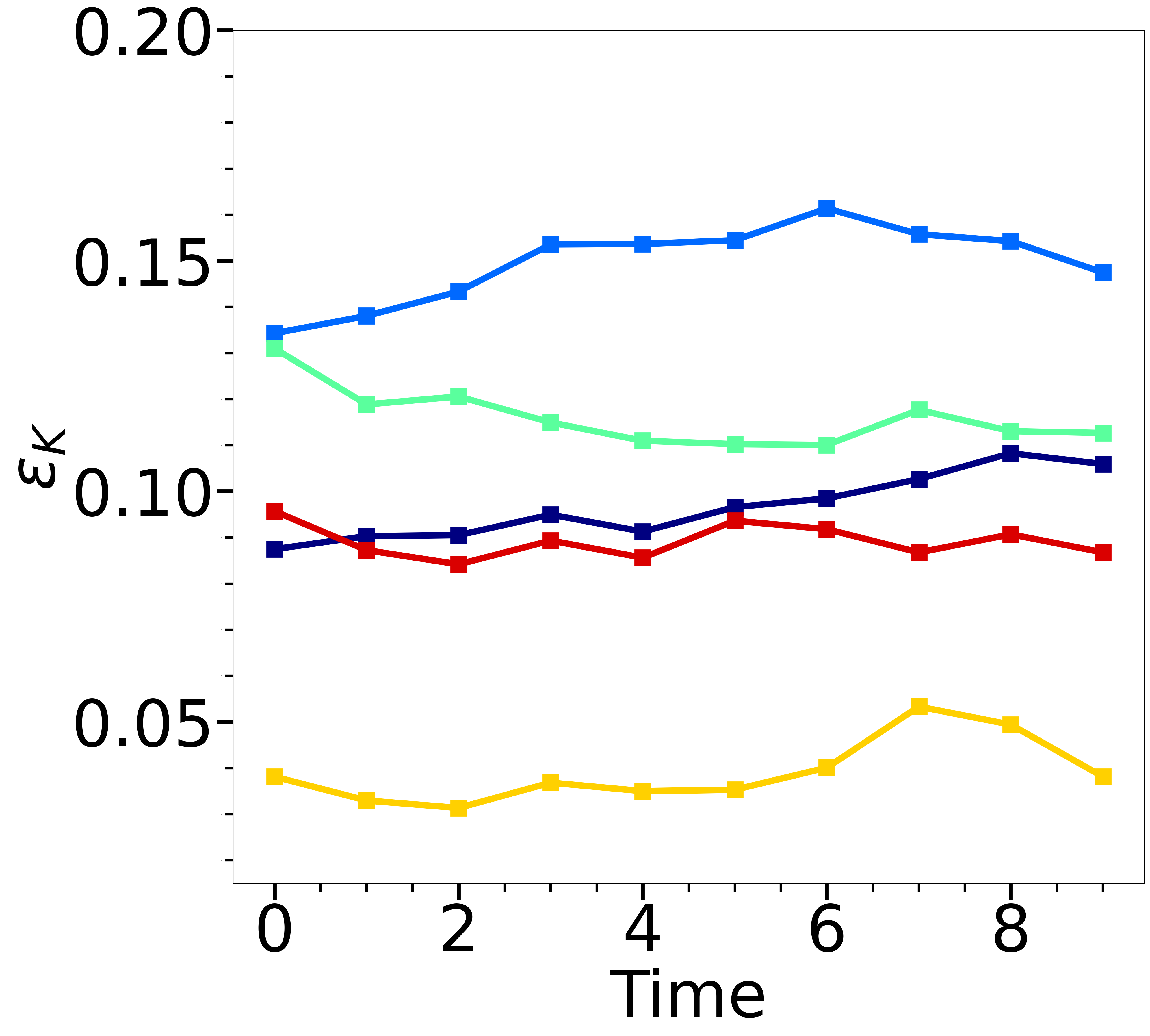}
\end{subfigure}
\begin{subfigure}
    \centering
    \includegraphics[width=0.24\textwidth,height=0.24\textheight,keepaspectratio]{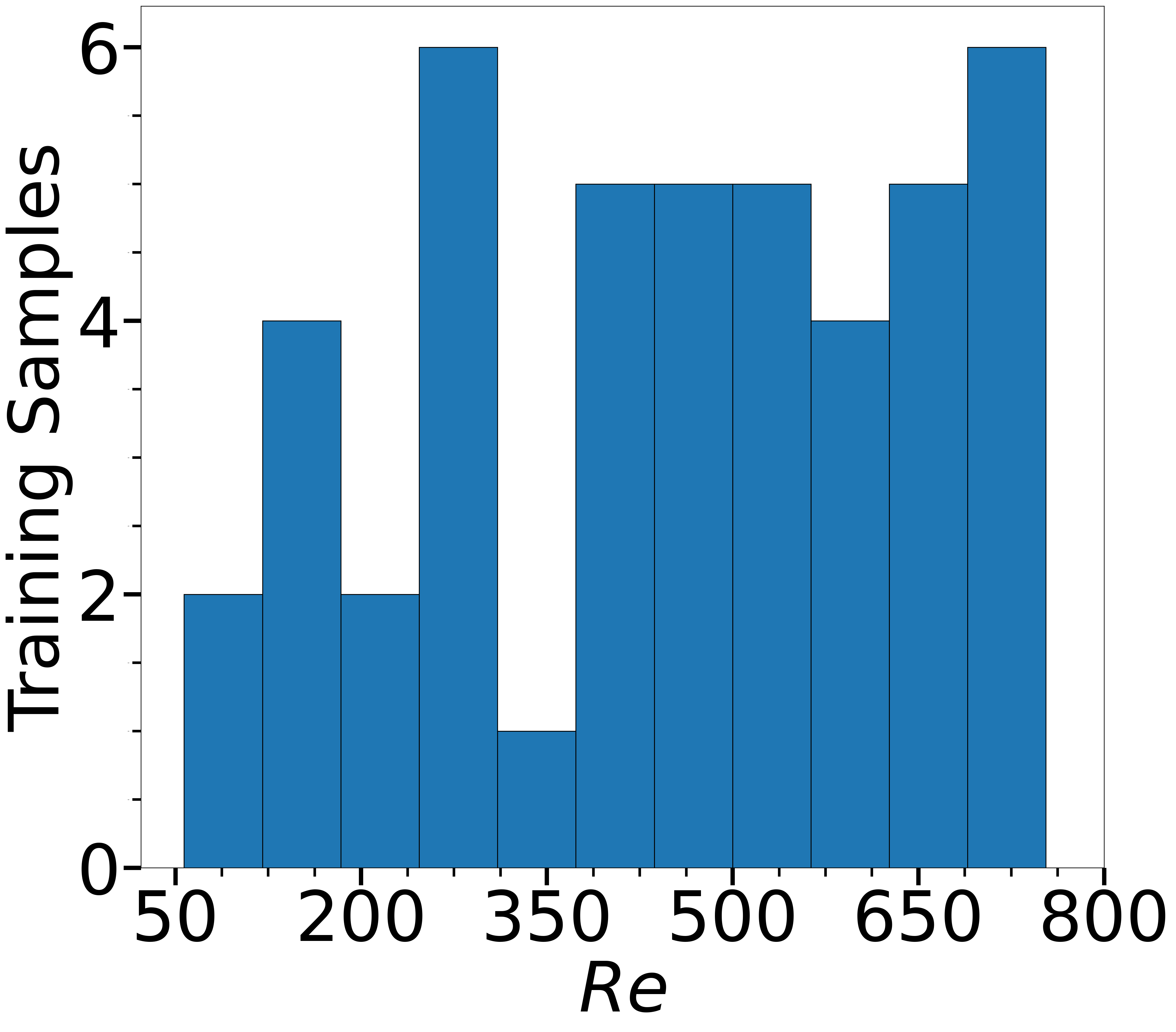}
\end{subfigure}
\caption{ Comparison of ${\varepsilon_E}$,${\varepsilon_{KE}}$, ${\varepsilon_K}$ on the 5 test conditions, along with a histogram of the 45 training conditions as a function of $Re$.  }
\label{fig: emulation_metrics}
\end{figure*}

\begin{figure*}[ht] 
\centering
\includegraphics[width= 0.95\textwidth]{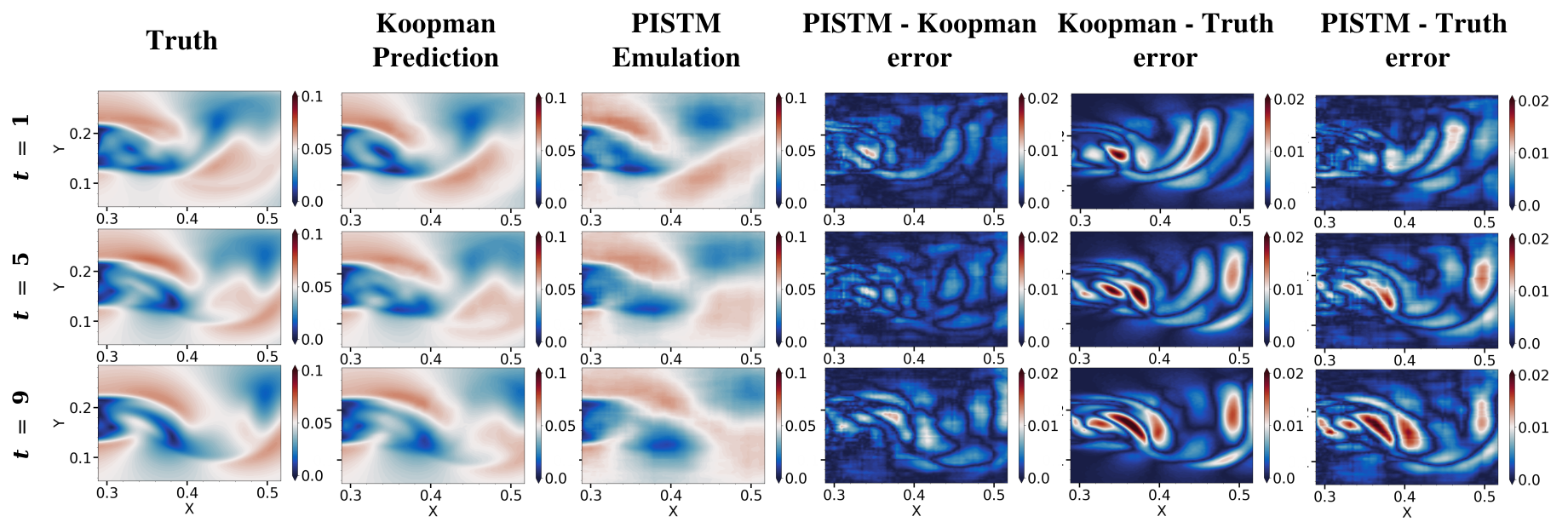} 
\caption{Qualitative comparison of the true data, Koopman predictions, emulated data, absolute Koopman emulation error, absolute Koopman error and absolute emulation error for the test case with $Re = 172$, at $t = 1, 5$ and $9$. }
\label{fig: emulation_image} 
\end{figure*}
\vspace{-2mm}
\section{Methodology}\label{sec: Methodology}
In order to solve this problem, we propose the following sequential approach:
\begin{enumerate}

\item Generate a Design of Experiments (DoE) in the space of operating conditions $\textbf{X}_{in}$. A part of the DoE, $\textbf{X}_{train}$, is used to generate training data, which is the spatio-temporal simulation of the dynamical system $f(t)$ from $t = T - h$ to $t = T - 1$.   

\item Leverage Koopman autoencoders for predicting $f(t)$ from $t = T$ to $t = T + k$ on the training set. This amounts to learning a separate Koopman operator $\mathcal{K}$ for each training condition based on $f(t)$ from $t = T - h$ to $t = T - 1$, and using the learned $\mathcal{K}$ to forecast the spatio-temporal dynamics, which has been shown to be an effective tool that incorporates physics-based constraints and stability in predicting how a dynamical system would evolve in future time-steps~\cite{Rice_koopman, Azencot_koopman}.  

\item Learn a ROM of the Koopman predictions in the time window of $t = T$ to $t = T + k$, based on the the training operating conditions $\textbf{X}_{train}$. This step is data-driven, where the training data is the spatio-temporal field data predicted by the Koopman operators on the training set. We employ a convolutional autoencoder architecture for this purpose, similar to the approaches proposed in \citet{mondal2022multifidelity, Maulik_SWE_GP}. The ROM learned in this stage provides a reduced order representation of how the Koopman predictions evolve for the dynamical system over a range of training operating conditions $\textbf{X}_{train}$. The outcome of this stage is an \textit{encoder}, and a \textit{decoder}. The encoder model ($\mathcal{F}_{enc}$) maps the high dimensional data into a reduced order \textit{latent} space ($\mathcal{F}_{enc} : \mathbb{R}^{D_{in}} \to \mathbb{R}^{D_{code}}$), while the decoder model ($\mathcal{F}_{dec}$) learns to reconstruct the high dimensional data from the latent space representation ($\mathcal{F}_{dec} : \mathbb{R}^{D_{code}} \to \mathbb{R}^{D_{in}}$). $D_{code}$ represents the dimension of the latent space, often termed as the \textit{code}, with $D_{code} << D_{in}$.

\item Learn a regression model $\mathcal{R}$ which predicts the latent space coefficients ($\textbf{z}$) for the ROM of the Koopman predictions as a function of the operating conditions and the time-steps of interest. We employ Gaussian process (GP) regression~\cite{Rasmussen2005} for this task, particularly because GPs are well-suited in problems of data-scarcity, which is a practical concern since the amount of training data in $\textbf{X}_{train}$ is expected to be limited due to expensive simulations. The outcome from this stage is a probabilistic latent space interpolation model denoted by $\mathcal{R} : (\textbf{X}_{in}, t) \mapsto \textbf{z}$.

\item For an unknown test point $\textbf{X}_{test} \in \textbf{X}_{in}$, the previous step provides the estimates of  $\Hat{\mathbf{z}}(t)$. These estimates can then be used as inputs to the pre-trained decoder $\mathcal{F}_{dec}$, to predict how the spatio-temporal dynamics evolve for the unknown test point int he time window of interest. 
\end{enumerate}
\begin{rem}    
It is to be noted that the problem statement can also be approached in a different way, whereby the spatio-temporal dynamics of an unknown operating condition from $t = T - h$ to $t = T - 1$ is first predicted using a spatio-temporal surrogate modeling framework, as demonstrated in~\citet{mondal2022multifidelity, Maulik_SWE_GP}, and then a Koopman operator is learned on the emulated data to predict the evolution in the future time-steps. However we avoid this approach for two practical reasons : (a) learning a Koopman operator during testing phase can be time-consuming, and become a bottleneck in the design process, and (b) emulation inaccuracies due to training data limitations can affect the training of the Koopman operator for an unknown conditions. Training data limitation is still a challenge in our proposed approach in the steps 3 and 4. However, with a sufficient amount of data, there is a scope to learn a data-driven model which predicts the Koopman dynamics as a function of operating conditions, which can be used to have fast and real-time predictions of the spatio-temporal evolution in the time window of $t = T$ to $t = T + k$ for any test condition in the domain of operating conditions. 
\end{rem}

We demonstrate our approach on a prototypical fluid flow problem : two-dimensional incompressible viscous flow around a circular cylinder~\cite{greco_paolillo_astarita_cardone_2020}. The equations are solved using the Lattice Boltzmann method (LBM), which is a popularly used technique in computational fluid dynamics to predict fluid flows~\cite{LBM_book}. The training set consists of 181 snapshots (each snapshot consisting of 80 $\times$ 80 spatial grid points)  of the absolute velocity fields for 45 Reynolds number ($Re$) values generated using Latin Hypercube Sampling (LHS) in the space of $50 < Re < 800$. These 181 snapshots correspond to the temporal history $t = T - h, \dots, T - 2, T - 1$ for each training case. Koopman autoencoders are learned for each of these training conditions, which predict the velocity fields for 10 timesteps in the future ($t = T, T + 1,\dots, T + k$). For simplicity of notation, values of $T = 0$, $k = 9$ and $h = 181$ are used, hereafter. The Koopman predicted velocity field snapshots are used to train the spatio-temporal emulation framework which includes the convolutional autoencoder and the Gaussian process regressors. This results in a spatio-temporal model which predicts the velocity fields from $t = 0, 1,\dots, 9$ for any input condition with $50 < Re < 800$. The results have been tested on a held out test set of 5 $Re$ conditions, $Re = 83, 172, 218, 406$, and $594$, which is completely unseen in the framework training phase. 
\vspace{-2mm}
\section{Results and Discussions} \label{sec: results}

For quantitative analysis of the prediction performance of the proposed PISTM framework, three different error metrics are considered from $t = 0, 1,\dots, 9$: ${\varepsilon_E}$ (relative prediction error between true and emulated data), ${\varepsilon_{KE}}$ (relative prediction error between Koopman prediction and emulated data) and ${\varepsilon_{K}}$ (relative prediction error between true data and Koopman prediction).The results are shown in Figure~\ref{fig: emulation_metrics}. Here, true data refers to the ground truth velocity field at the test conditions obtained from simulation, emulated data refers to the prediction from our proposed PISTM framework, and Koopman predictions refer to the predictions on the test conditions provided by Koopman operators trained on the true data. ${\varepsilon_{K}}$ is an indicator of how the Koopman operator predictions compare with the truth, when trained on the true historical data of the test conditions. On the other hand, ${\varepsilon_{KE}}$ is an indicator of how our PISTM predicted spatio-temporal evolution compare with the predictions of Koopman operators trained on true historical data. It is reiterated that the PISTM framework does not observe the true historical data for a test condition. It is seen that ${\varepsilon_{KE}}$ remains approximately constant over the prediction horizon for each test condition, which is an outcome of the physics-constrained stability criterion enforced in the formulation of the Koopman autoencoders~\cite{Azencot_koopman, Rice_koopman}. This has also been observed with respect to the cylinder flow problem by~\citet{Erichson_Lyapunov}, whereby it was seen that enforcing constraints based on the underlying dynamical system can result in stable forecasting performance, whereas the forecasting error from physics-agnostic models typically diverge with an increase in time. Using the proposed PISTM framework, it is possible to have quite accurate predictions of the Koopman evolution at the test conditions, which is reflected in ${\varepsilon_{KE}} < 0.10$ for all test cases except $Re = 83$. The higher ${\varepsilon_{KE}}$ for $Re = 83$ can be attributed to the training data distribution, whereby only 2 training samples are present in the range of $50 < Re < 100$. Nevertheless, with a sufficiently high number of training conditions, ${\varepsilon_{KE}}$ is expected to be low across $\textbf{X}_{in}$. The variation of ${\varepsilon_{E}}$ suggests that the proposed PISTM framework yields stable and accurate (comparable to the accuracy of the Koopman predictions) estimates of how the system evolves for unknown test conditions over a temporal horizon beyond the scope of the training data.  
Figure~\ref{fig: emulation_image} shows a qualitative comparison of the true data, Koopman predictions, PISTM emulated data, absolute error between PISTM and Koopman predictions, absolute error between Koopman prediction and truth, and absolute error between PISTM emulation and truth for the test case with $Re = 172$, at the time-steps $t = 1, 5$ and $9$. It is seen that the emulated flowfields are in close agreement with the truth, with the spatial distribution of the absolute emulation error closely matching that of the absolute Koopman prediction error. The predictions are near real-time for the unknown test conditions, with the spatio-temporal evolution for each test case being predicted in$\sim$ 3s, as compared to $\sim$ 170 mins required to perform the actual simulations, thus providing$~\sim \mathcal{O}(10^3)$x speedup in the testing phase. Moreover, this approach also does not involve learning a Koopman autoencoder in the testing phase, which itself is time-consuming (learning each Koopman autoencoder from the training conditions take $\sim$30 mins). 
This shows that, by combining data-driven spatio-temporal modeling with the Koopman predictions, it is possible to obtain fast and accurate and physics-constrained forecasts for a dynamical system at unseen operating conditions. This can be particularly helpful for tackling the computational bottleneck  in practical engineering design problems, where the underlying simulations of the dynamical systems are expensive. 
\vspace{-3mm}
\section{Conclusions}
In this paper we present a novel physics-informed spatio-temporal modeling (PISTM) framework which can result in fast and accurate, physics-constrained predictions of a dynamical system for unseen operating conditions over a specified temporal horizon beyond the scope of the training observations, which results in low generalization errors. The key aspect of this framework is its non-intrusiveness, i.e. it does not require any knowledge of underlying physical equations in its formulation. For engineering design problems where the cost of generating data from an underlying dynamical system at unknown operating conditions is high, the proposed framework can significantly accelerate the overall process. Although the the performance of this framework has been demonstrated on a fluid flow problem in this work, it can be potentially applied for emulating the spatio-temporal output of any nonlinear dynamical system.
\vspace{-4mm}
\section*{Acknowledgments}
This material is based upon work supported by internal funding of Raytheon Technologies Research Center. 

{\scriptsize
\bibliography{ref_bib}}
\bibliographystyle{icml2025}


\end{document}